\documentclass[10pt,twocolumn,letterpaper]{article}

\usepackage[pagenumbers]{cvpr} % To force page numbers, e.g. for an arXiv version

\usepackage{graphicx}
\usepackage{amsmath}
\usepackage{amssymb}
\usepackage{booktabs}
\usepackage{multicol}
\usepackage{multirow}
\usepackage{makecell}
\usepackage{algorithm}
\usepackage{algpseudocode}
\usepackage[table]{xcolor}
\usepackage{comment}
\usepackage{tabularx}

\def\mathbi#1{\textbf{\em #1}}

\usepackage[pagebackref,breaklinks,colorlinks]{hyperref}

\usepackage[capitalize]{cleveref}
\crefname{section}{Sec.}{Secs.}
\Crefname{section}{Section}{Sections}
\Crefname{table}{Table}{Tables}
\crefname{table}{Tab.}{Tabs.}

\algblockdefx{FORALLP}{ENDFAP}[1]%
  {\textbf{for all }#1 \textbf{do in parallel}}%
  {\textbf{end for}}
\algblockdefx{FUNCDO}{ENDFUNCDO}[1]%
  {#1 \textbf{do}}%
  {\textbf{end}}
\DeclareMathAlphabet\mathbfcal{OMS}{cmsy}{b}{n}

\def\cvprPaperID{***} % *** Enter the CVPR Paper ID here
\def\confName{CVPR}
\def\confYear{2022}

\begin{document}

\title{MeMOT: Multi-Object Tracking with Memory}

\author{
    Jiarui Cai$^1$\thanks{The work was done during an Amazon internship.} \hspace{0.35cm} Mingze Xu$^2$\thanks{Corresponding Author.} \hspace{0.25cm} Wei Li$^2$ \hspace{0.25cm} Yuanjun Xiong$^2$ \hspace{0.25cm} Wei Xia$^2$ \hspace{0.25cm} Zhuowen Tu$^2$ \hspace{0.25cm} Stefano Soatto$^2$ \\ [.5ex] $^1$University of Washington \hspace{0.9cm} $^2$AWS AI Labs \\ [.5ex]
    {\tt\small jrcai@uw.edu, \{xumingze,wayl,yuanjx,wxia,ztu,soattos\}@amazon.com}
}

\maketitle

%%%%%%%%% ABSTRACT
\begin{abstract}
    We propose an online tracking algorithm that performs the object detection and data association under a common framework, capable of linking objects after a long time span. This is realized by preserving a large spatio-temporal memory to store the identity embeddings of the tracked objects, and by adaptively referencing and aggregating useful information from the memory as needed. Our model, called MeMOT, consists of three main modules that are all Transformer-based: 1) Hypothesis Generation that produce object proposals in the current video frame; 2) Memory Encoding that extracts the core information from the memory for each tracked object; and 3) Memory Decoding that solves the object detection and data association tasks simultaneously for multi-object tracking. When evaluated on widely adopted MOT benchmark datasets, MeMOT observes very competitive performance.
\end{abstract}

%%%%%%%%% BODY TEXT
\section{Introduction}
\label{sec:intro}

Online multi-object tracking (MOT)~\cite{wojke2017simple,feichtenhofer2017detect,bergmann2019tracking,zhang2020fair} aims at localizing a set of objects (\eg, pedestrians), while following their trajectories over time so that the same object bears the same identities in the entire input video stream. 
Earlier methods mostly solved this problem with two separate stages:
1) the object detection stage that detects object instances in individual frames~\cite{felzenszwalb2009object,ren2015faster,liu2016ssd,zhou2019objects,ge2021yolox}; and 2) the data association stage that links the detected object instances across time~\cite{braso2020learning,zhang2020fair} by modeling the state changes of tracked objects and solving a matching problem between them and the detection results.
Though recent studies~\cite{meinhardt2021trackformer,zeng2021motr} suggest that combining these two stages could be beneficial, the combination usually leads to the undesired simplification of the association module in modeling the change of the objects in time.  

\begin{figure}
    \centering
    \includegraphics[width=0.97\linewidth]{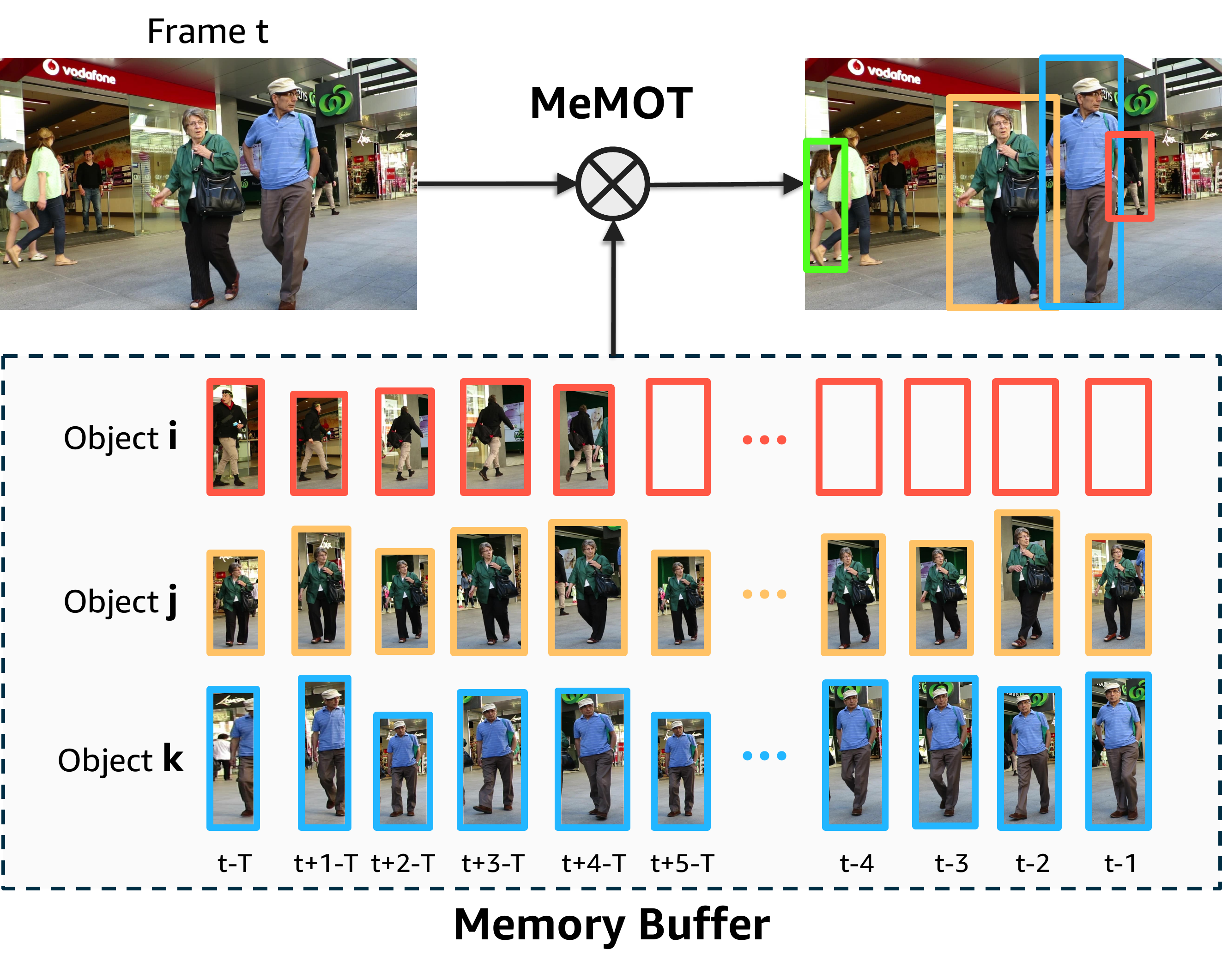}  
    \vspace{-4.5mm}
    \caption{\textbf{Illustration of the idea of MeMOT}.
    A spatio-temporal memory stores a long range states of all tracked objects and is updated over time.
    Each row in the memory buffer represents an active tracklet.
    The ``person crops'' indicate that their the history states are preserved in the memory, and the blank box indicates this person does not appear in the frame at that time, occluded or not detected.
    The tracking plots show that MeMOT can maintain active tracks (yellow and blue boxes), link reappearing tracks after occlusion (red box), and generate new objects (green box).}
    \vspace{-2.5mm}
    \label{fig:teaser}
\end{figure}

% It requires not only accurate object detection in individual video frames, but also robust association~\cite{braso2020learning,zhang2020fair} to link objects using their ever-changing appearances and positions.
In this paper, we propose a Transformer-based tracking model, called \emph{MeMOT}, that performs object detection and association under a common framework in an online manner. 
The key design of MeMOT is to build a large spatio-temporal memory that stores the past observations of the tracked objects. The memory is actively encoded in every time step by referencing relevant information so that the states of the objects are more accurately approximated for the association task.
The rich representation of the tracked objects extracted from the spatial-temporal memory enables us to solve the object detection and association tasks in a unified decoding module. It directly outputs object instances that have been tracked and reappears in the latest frame and novel object instances that are first time seen.
The idea of MeMOT is illustrated in Fig.~\ref{fig:teaser}. 

\begin{figure*}[t!]
    \centering
    \includegraphics[width=0.97\textwidth]{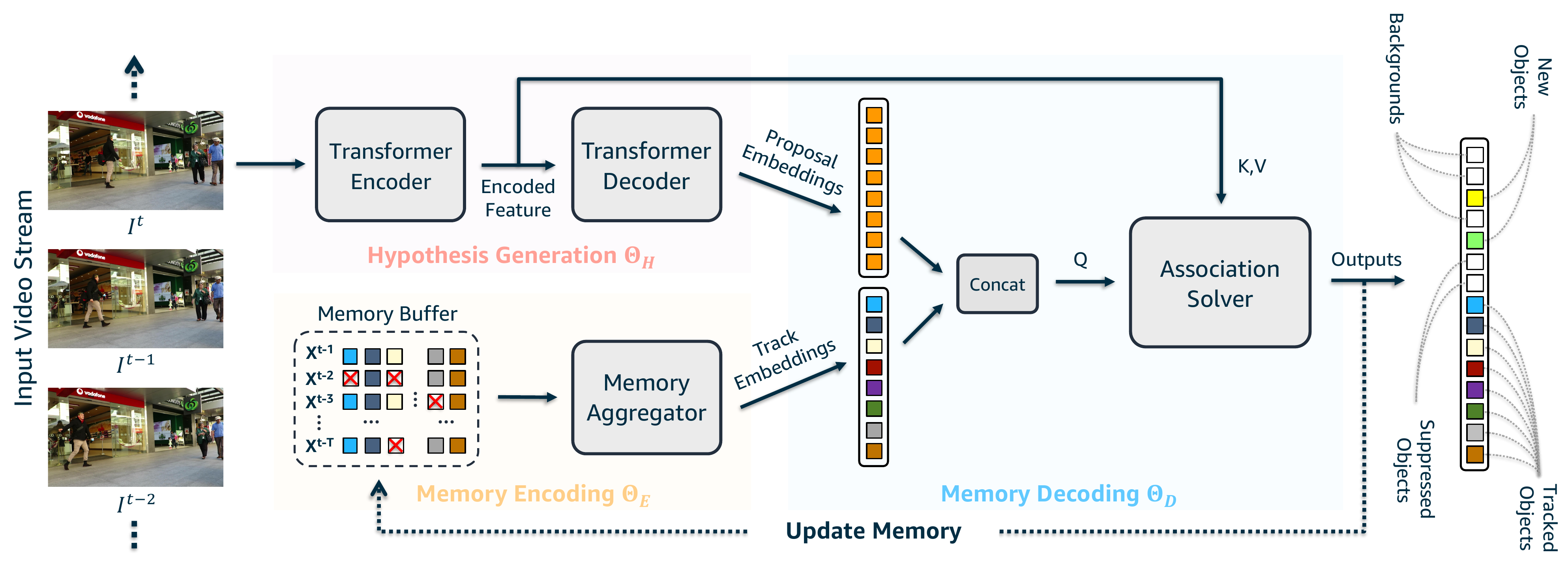}
    \vspace{-3.5mm}
    \caption{
        \textbf{Visualization of MeMOT}, which runs three main components: 1) a hypothesis generation module $\Theta_H$ that produces object proposals for the current video frame, 2) a memory encoding module $\Theta_E$ that retrieves core information for each tracked objects, and 3) a memory decoding module $\Theta_D$ that solves the object detection and data association tasks simultaneously.
        MeMOT maintains a memory buffer to store long-range states of tracked objects, together with an efficient encoding-decoding process that retrieves useful information for linking objects after a long time span.
        Each hypothetical object is predicted as a new object, a tracked object, or a background region.
    }
    \label{fig:network}
\end{figure*}

At each time step, MeMOT runs the following three main components:
1) a \textit{hypothesis generation} module that produces object proposals from input image feature maps as a set of embedding vectors;
2) a \textit{memory encoding} module that encodes the spatial-temporal memory corresponding to each tracked object into a vector known as the track embedding;
and 3) a \textit{memory decoding} that inputs the proposal and track embeddings and solves the object detection and data association tasks simultaneously for multi-object tracking.
The hypothesis generation module is implemented by a Transformer-based encoder-decoder network~\cite{carion2020end,zhu2020deformable}.
It produces a set of embedding vectors, known as the \emph{proposal embedding}, each representing one hypothetical object instance. 
The memory encoding module first divides the spatial temporal memory on each object into short- and long-term memories and aggregates them each into one embedding vector through cross attention modules~\cite{vaswani2017attention}. The two vectors then interact through the self-attention mechanism to produce the \emph{track embedding} of the tracked object at this time step. 
The proposal and track embeddings, together with the original image features, are then fed to the memory decoding module.
For each track embedding, it produces the location and the visibility of the object being tracked in this frame. For each proposal embedding, it predicts whether this hypothetical object instance is depicting a novel object, a tracked object, or simply a background region. 
An illustration of the MeMOT model is shown in Fig.~\ref{fig:network}.
The entire model can be trained end to end on video datasets with object bounding box and identity annotations.
During inference, we obtain the tracking outputs in one inference run of the model at each time step, \emph{without} any extra optimization~\cite{rangesh2021trackmpnn, chu2021transmot} or post-processing~\cite{sun2020transtrack, bergmann2019tracking, zhang2020fair}. 

We evaluate MeMOT on the MOT Challenge~\cite{milan2016mot16,dendorfer2020mot20} benchmarks for pedestrian tracking.
Experimental results show that MeMOT achieves the state-of-the-art performance among all algorithms with an in-network association solver and is competitive with those utilizing a post-network association process.
Specifically, MeMOT outperforms other Transformer-based methods in both object detection and data association.
Extensive ablation studies further validate the design and effectiveness of MeMOT.

\section{Related Work}
\label{sec:related_work}

\noindent \textbf{Classical Tracking Methods}.
Tracking is well studied in computer vision~\cite{isard1998condensation,babenko2010robust,wu2013online,kristan2015visual}.
Coping with the underlying uncertainties of the tracking results~\cite{isard1998condensation} and object appearances/positions/shapes~\cite{babenko2010robust} has been a central challenge.
Classical non-deep learning methods~\cite{wu2013online} have laid out solid mathematical and statistical foundations.
Specifically, Kalman~\cite{welch1995introduction} and particle filters ~\cite{gustafsson2002particle} are widely adopted for tackling tracking problems~\cite{shen2003probabilistic,hue2001particle,xing2009multi}. The progressive observation-based Bayesian inference method~\cite{xing2010multiple} is proposed for MOT in online sports videos. A spatial and temporal shape representation-based Bayesian framework~\cite{giebel2004bayesian} is proposed for multi-cue 3D deformable object tracking. In these methods,
an optimal filter maintains tracking states that summarize history information and estimate new frame's tracking results. In a linear-Gaussian case, the optimal state can indeed be estimated, while for more general non-linear, non-Gaussian cases, it is difficult to estimate the optimal state with a finite-dimensional state representation. For instance, occlusion in visual multiple object tracking is clearly non-linear and non-Gaussian.
To tackle this challenge, tracking methods~\cite{choi2010multiple,perera2006multi} that can access multiple frames states (offline tracking) is desired.

\vspace{2pt} \noindent \textbf{MOT with CNNs}.
A typical scheme for MOT~\cite{wojke2017simple,feng2019multi,wang2019exploit,chu2019famnet} is ``tracking-by-detection'', which directly uses ready-made detectors~\cite{felzenszwalb2009object,ren2015faster,liu2016ssd,zhou2019objects,ge2021yolox} and focuses on the data association.
% Recent approaches build object detection and data association in the same network.
Tracktor++~\cite{bergmann2019tracking} propagates the bounding boxes of tracked objects as region proposals to the next frame.
CenterTrack~\cite{zhou2020tracking} takes an additional point-based heatmap as input and matches objects anywhere within the receptive field.
JDE~\cite{xu2018joint,wang2019towards,zhang2020fair,li2021semi} is built with two homogeneous branches for object detection and ReID feature extraction, respectively.
Joint detection and tracking models improve the runtime, but sacrifice the tracking recovery after occlusion and cannot reconnect long-term missing objects.

\vspace{2pt} \noindent \textbf{MOT with Transformers}.
Vision Transformers have been successfully applied in image recognition~\cite{carion2020end,dosovitskiy2020image,zhu2020deformable,liu2021swin} and video analysis~\cite{arnab2021vivit,bertasius2021space,sharir2021image,liu2021video} lately.
In tracking, TrackFormer~\cite{meinhardt2021trackformer} and MOTR~\cite{zeng2021motr} simultaneously perform the object detection and association by concatenating the object and autoregressive
track queries as inputs to the Transformer decoder in the next time step.
On the other hand, TransCenter~\cite{xu2021transcenter} and TransTrack~\cite{sun2020transtrack} only use Transformers as feature extractor and recurrently pass track features to aggressively learn aggregated embedding of each object.
TransMOT~\cite{chu2021transmot} still uses CNNs as detector and feature extractor, and learns an affinity matrix with Transformers.
The above work explores the mechanism of representing object states as dynamic embeddings. However, the modeling of long-term spatio-temporal observations and adaptive feature aggregation methods are underdeveloped.

\vspace{3pt} \noindent \textbf{Memory Networks}.
Pioneering work using memory networks has been proposed in NLP~\cite{graves2014neural,weston2014memory,sukhbaatar2015end} by focusing on temporal reasoning tasks such as question answering~\cite{xiong2016dynamic,kumar2016ask} and dialogue systems~\cite{wu2019global}.
Video analysis tasks, such as action recognition~\cite{wu2019long,xu2021long}, and video object segmentation~\cite{oh2019video,lu2020video}, leverage an external memory to store and access time-indexed features in prolonged sequences, significantly improving the ability to remember the past.
Recently, memory networks have been introduced into tracking.
MemTrack~\cite{yang2018learning} reads a residual template from memory and combines it with the initial template to update the representations of targets.
STMTrack~\cite{fu2021stmtrack} guides the information retrieval with the current frame and adaptively obtains all useful information as it needs.
However, these work focuses on single object tracking (SOT), and does not need to concern about the inter-object association.
We propose to use a large spatio-temporal memory to achieve robust object association across time for MOT.

% \section{The Proposed Method}
\section{Multi-Object Tracking with Memory}
\label{sec:method}

\subsection{Overview}
\label{sec:method:overview}

Given a sequence of video frames $\mathbi{I} = \{I^0, I^1, \cdots, I^T\}$,
the goal of online MOT is to localize a set of $K$ objects while following their trajectories $\mathbfcal{T} = \{\mathcal{T}_0, \mathcal{T}_1, \cdots, \mathcal{T}_K\}$ over time by causal processing.
In this paper, we propose an end-to-end tracking algorithm, called \textit{MeMOT}, which jointly learns the object detection and association.
Different from most existing methods~\cite{bergmann2019tracking} that propagate the states of tracked objects between adjacent frames,
we build a \textit{spatio-temporal memory} that stores long-range states of all tracked objects, together with a \textit{memory encoding-decoding} process that efficiently retrieves useful information for linking objects after a long time span.

Specifically, as shown in Figure~\ref{fig:network}, MeMOT consists of three main components:
1) a frame-level hypothesis generation module $\Theta_H$ that produces region proposals for the current video frame $I^t$,
2) a track-level memory encoding module $\Theta_E$ that aggregates track embeddings,
and 3) a memory decoding module $\Theta_D$ that associates the new detections with tracked objects.
At time step $t$,
$\Theta_H$ generates $N^t_{pro}$ region proposals, represented as proposal embeddings $\mathbi{Q}_{pro}^t \in \mathbb{R}^{N^t_{pro} \times d}$ using a Transformer-based architecture.
The memory encoder $\Theta_E$ adaptively translates the ``history states'' of each track into one compact representation, denoted as track embeddings $\mathbi{Q}_{tck}^t \in \mathbb{R}^{N^t_{tck} \times d}$. 
By querying the encoded image feature with $[\mathbi{Q}_{pro}^t, \mathbi{Q}_{tck}^t]$,
the memory decoder $\Theta_D$ computes the inter-object relations and updates the embeddings as $[\widehat{\mathbi{Q}}_{pro}^t, \widehat{\mathbi{Q}}_{tck}^t]$ accordingly.
Then the locations $[\mathbi{B}_{pro}^t, \mathbi{B}_{tck}^t]$ and confidence scores  $[\mathbi{S}_{pro}^t, \mathbi{S}_{tck}^t]$ of new and tracked objects are predicted based on these output embeddings.
Finally, the locations and states of the previously tracked objects are used to update their trajectory and the memory.
The ``new-born'' objects are initialized in $\mathbfcal{T}$ and their states are added into the memory.

\subsection{Hypothesis Generation}
\label{sec:method:theta_H}

The hypothesis generation network $\Theta_H$ is built with an encoder-decoder architecture based on Transformers~\cite{carion2020end,zhu2020deformable}.
It produces a set of $N^t_{pro}$ region proposals that either initiate ``new-born'' objects for the current video frame or update tracked objects with their latest identity and position information.
The $\Theta_H$ encoder takes a sequentialized feature map $z_0^t \in \mathbb{R}^{C \times HW}$ as inputs, which is extracted by a CNN backbone from the input frame $I^t$.
Each element in $z_0^t$ is supplemented with a unique positional encoding that indicates its spatial location.
The image feature is encoded as $z_1^t \in \mathbb{R}^{d \times HW}$ using a multi-layer Transformer encoder.
The $\Theta_H$ decoder receives the encoded feature $z_1^t$ and empty object queries (represented as learnable embeddings), and produces the final set of proposal embeddings $\mathbi{Q}_{pro}^t \in \mathbb{R}^{N_{pro}^t \times d} $.
The objectness scores and bounding boxes of each proposal can be predicted from $\mathbi{Q}_{pro}^t$.

\begin{figure}
    \centering
    \includegraphics[width=0.5\textwidth]{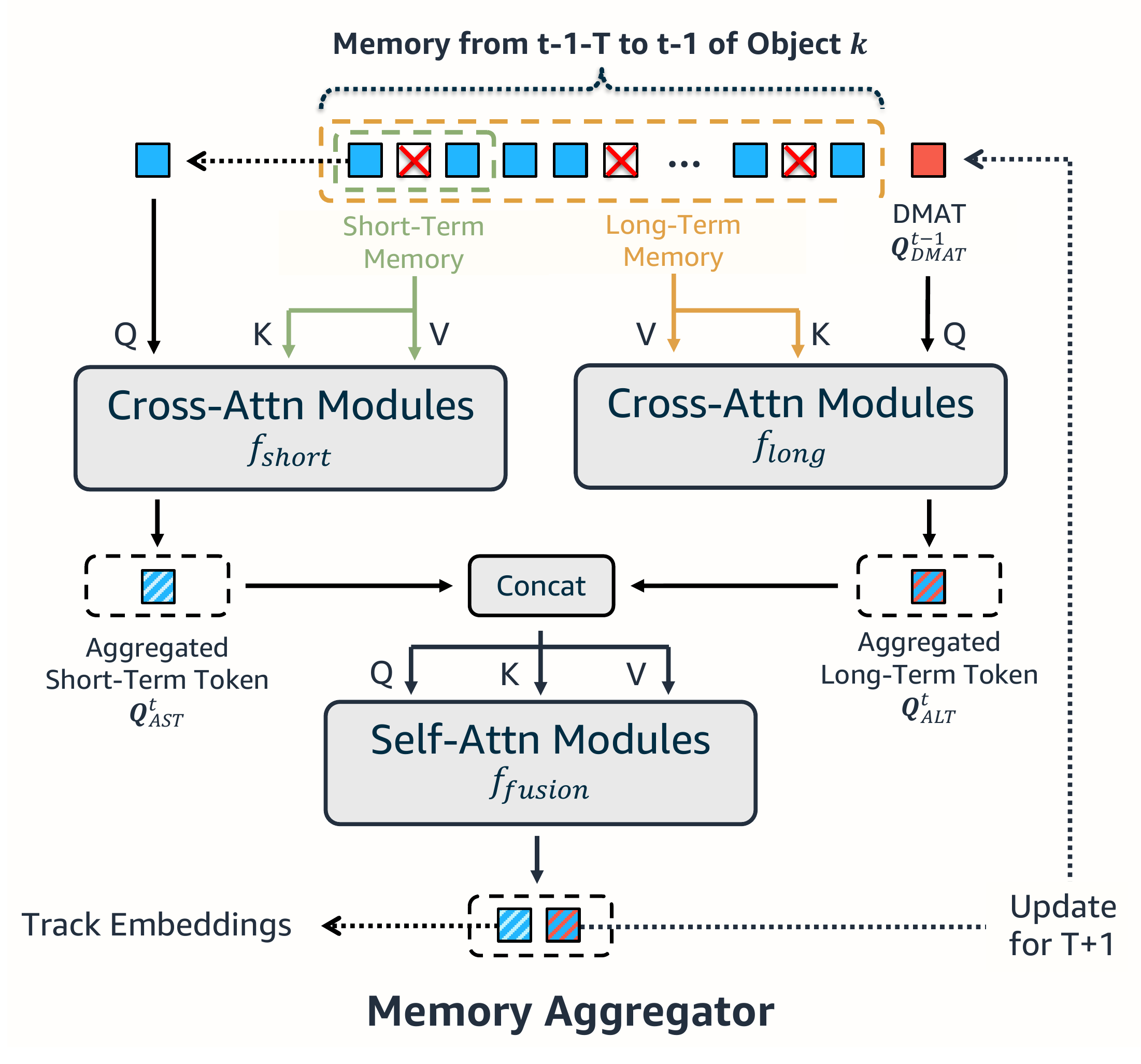}
    \vspace{-7.5mm}
    \caption{
        \textbf{Illustration of Memory Aggregator}, which consists three attention modules: 1) short-term $f_{short}$ that smoothes out noises in recent frames, 2) long-term $f_{long}$ that extracts supportive features from long-range context, and 3) fusion blocks that aggregate short- and long-term branches. The aggregated embeddings will be used as the track embeddings (blue-white query) and update DMAT (blue-red query) for the next time step.
        % \textbf{Illustration of Memory Aggregator}, which consists three attention modules: 1) a short-term $f_{short}$, 2) a long-term $f_{long}$, and 3) a fusion block. The aggregated embeddings will be used as the track embeddings (blue-white query) and update DMAT (blue-red query) for the next time step.
    }
    \label{fig:ta}
    \vspace{-4.5mm}
\end{figure}

\subsection{Spatio-Temporal Memory}
\label{sec:method:memory}

We store the history states of all $N$ tracked objects in a spatio-temporal memory buffer $\mathbi{X} \in \mathbb{R}^{N \times T \times d}$.
It reserves at most $N_{max}$ objects and a maximum of $T_{max}$ time steps for each object.
The memory is implemented with a first-in-first-out (FIFO) data structure. 
At time step $t$, the memory is represented as the states of $N_{tck}^{t-1}$ active objects in the past $T$ frames, $\mathbi{X}^{t-1-T:t-1}=\{x_k^{t-1-T:t-1}\}_{k=1:N_{tck}^{t-1}}$, where $x_k^{t-1-T:t-1}$ is the states of the $k$-th object and is padded with $\mathbi{0}$ if this object does not appear in the frame $I^t$.
When $T$ is larger than $T_{max}$, the ``oldest'' state $x_k^{t-1-T}$ of each tracklet graduates from the memory.
$N_{max}$ is set to be significantly large (\eg, 300 or 600) to cover the typical number of objects in a video, and a choice of $T_{max}$ is 24.

\subsection{Memory Encoding}
\label{sec:method:theta_E}

As shown in Fig.~\ref{fig:ta}, we encode the memory and extract the track embedding with three attention modules:
1) a short-term block $f_{short}$ for assembling embeddings of neighboring frames to smooth out the noises,
2) a long-term block $f_{long}$ for extracting relevant features in the temporal window covered by the memory,
and 3) a fusion block $f_{fusion}$ for aggregating embeddings from short- and long-term branches.

For each tracklet, the short-term module $f_{short}$ takes as inputs its previous $T_s$ states while the long-term memory module $f_{long}$ utilizes longer history with length of $T_l$ ($T_s \ll T_l$).
$f_{short}$ and $f_{long}$ are implemented with multi-head cross-attention modules, where the history states are key and value inputs.
% We assume that the observations in the current frame are closer to those in the nearby frames and the long-term memory contains more redundant but supportive information.
The input query for $f_{short}$ is the most recent state $\mathbi{X}^{t-1}$, while an dynamically updated embedding, called \textit{Dynamic Memory Aggregation Tokens (DMAT)}, $\mathbi{Q}^{t-1}_{dmat} = \{{q_{dmat}}_k^{t-1}\}_{k=1:N_{tck}}$ is used for $f_{long}$.
When every tracklet is initiated, it is associated with the same DMAT as others; after that, at time step $t > 0$, DMAT is iteratively updated from the previous step. This design will be further validated in Sec.~\ref{sec:exp:ablation}.
The outputs of the short- and long-term branches, denoted as \textit{Aggregated Short-term Token (AST)} $\mathbi{Q}^t_{AST}$ and \textit{Aggregated Long-term Token (ALT)} $\mathbi{Q}^t_{ALT}$, are then fused by $f_{fusion}$.
It outputs the track embedding $\mathbi{Q}_{tck}^t$ and an updated $\mathbi{Q}^{t}_{dmat}$ where the latter is retained for the next timestep. 

\subsection{Memory Decoding}
\label{sec:method:theta_D}

The memory decoder $\Theta_D$ takes the proposal embedding, track embedding, and the image feature as inputs to produce the final tracking results. 
It is realized by using stacked Transformer decoder units, where the concatenated proposal and track embeddings $[\mathbi{Q}^t_{pro}, \mathbi{Q}^t_{tck}]$ are used as queries.
$\Theta_D$ takes the encoded image feature $z_1^t$ from $\Theta_H$ as key and value.

For each entry $q_i^t$ in $\Theta_D$'s outputs $[\widehat{\mathbi{Q}}^t_{pro}, \widehat{\mathbi{Q}}^t_{tck}]$, the decoding process generates three predictions: the bounding box (in the format of offsets \wrt the learned reference points), the \emph{objectness score}, and the \emph{uniqueness score}.
The objectness score $o_{i}^t$ for a query $q_i^t$ ranges from $0$ to $1$, where $o_{i}^t=1$ means the model determines the entry is depicting a visible object.
The uniqueness score $u_i^t$ also ranges from $0$ to $1$. When $u_i^t=1$ the model predicts that the object depicted by $q_i^t$ is unique and should be included in the tracking outputs. Otherwise it needs to be suppressed. We define that $u_i^t=1$ if $q_i^t\in \widehat{\mathbi{Q}}^t_{tck} $. When the model learns to predict $u_i^t$ for each proposal entry, we enforce that a proposal is only considered novel ($u_i^t=1$) when it is not related to any object being tracked. 
We can then define a unified \textbf{confidence score} for both proposal and track entries as the multiplication of objectiveness and uniqueness scores:
% as $s_k^t = o_k^t \cdot u_k^t$:
%
\vspace{-1.0mm}
\begin{equation}
s_k^t = 
    o_k^t \cdot u_k^t.
\label{eq:score}
\end{equation}
\vspace{-1.0mm}
The predicted confidence scores of the proposal and track queries are referred to as $\mathbi{S}^t_{tck}$ and $\mathbi{S}^t_{pro}$, respectively. 
For each entry $q_i^t$, the model predicts its bounding box $\mathbf{b}_i^t$, where $\mathbf{b}_i^t \in \mathbb{R}^{4 \times 1}$ includes the object's center coordinates, width, and height. 

The above formulation allows us to solve the object detection and data association problem simultaneously.
In inference, we threshold each entry of $[\widehat{\mathbi{Q}}^t_{pro}, \widehat{\mathbi{Q}}^t_{tck}]$ with the threshold $\epsilon$ and only retain entries with $s_i^t \ge \epsilon$. The resulted entries will automatically bear an track identity or initialize a new track according to whether they come from  $\widehat{\mathbi{Q}}^t_{pro}$ or $\widehat{\mathbi{Q}}^t_{tck}$.
We can then obtain the final tracking results by combining the inherited or newly formed track identities with the corresponding bounding box prediction. There is \emph{no} need for further post processing~\cite{wojke2017simple,bergmann2019tracking,zhang2020fair}. 

\begin{figure}
    \centering
    \includegraphics[width=0.5\textwidth]{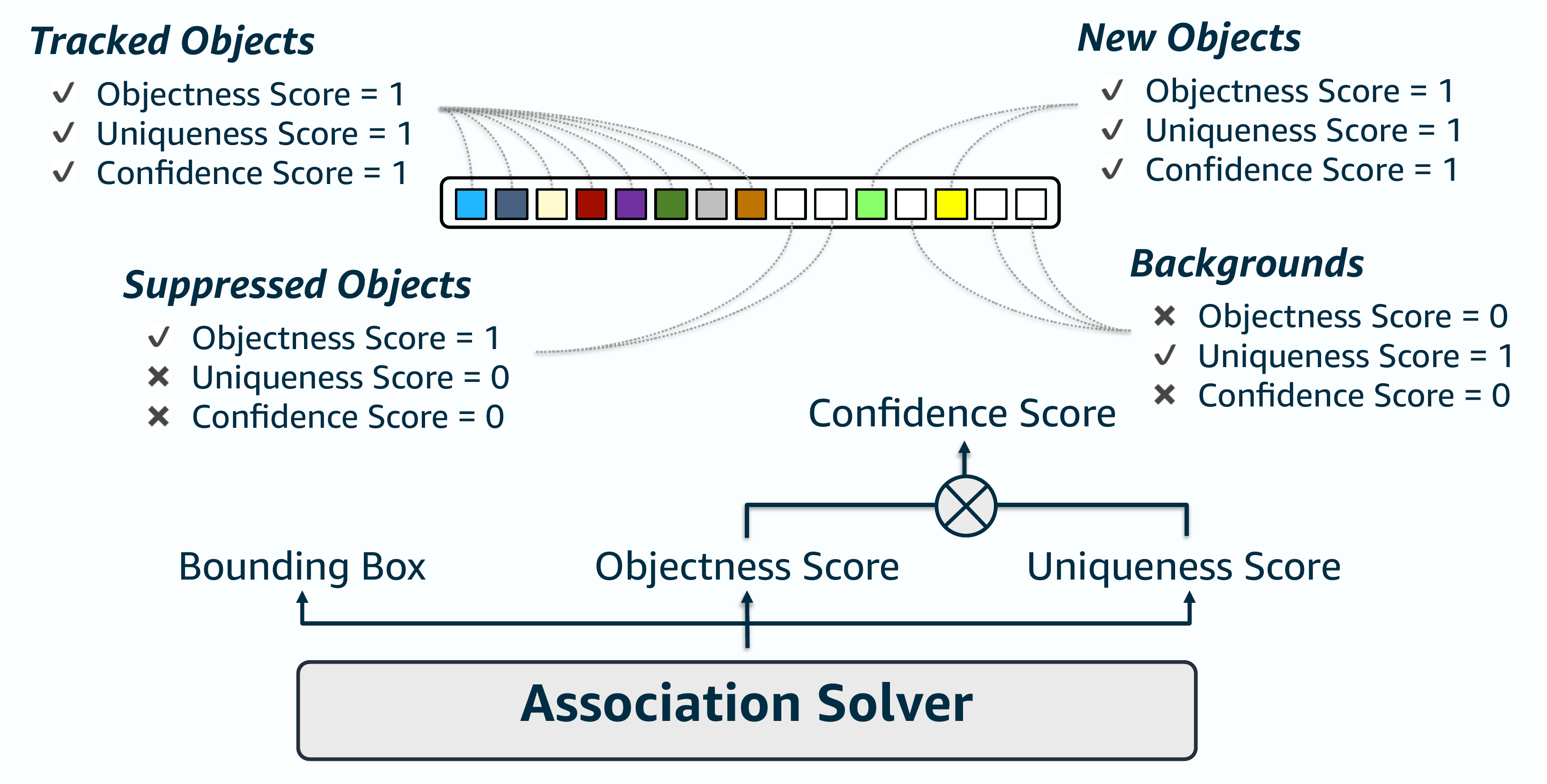}
    \vspace{-6.0mm}
    \caption{\textbf{Illustration of ground truth assignment} to tracked objects, new objects, suppressed objects, and backgrounds. We show the assigned groundtruth scores for each type of entry.}
    \label{fig:asso_solver}
    \vspace{-4.0mm}
\end{figure}

To generate the supervision signals for $o_i^t$, $u_i^t$, and $\mathbf{b}_i^t$ on each frame, we first assign the objectiveness score and bounding box to entries in $\widehat{\mathbi{Q}}^t_{tck}$ based on whether the tracked object is present in this frame. 
Then for each entry in $\widehat{\mathbi{Q}}^t_{pro}$, we assign the groundtruth bounding boxes, regardless of new or already tracked, to each entry through bipartite matching~\cite{erhan2014scalable,carion2020end}.
Then we assign the groundtruth uniqueness score to each proposal entry, as shown in Fig.~\ref{fig:asso_solver}, based on whether its matched object has been seen before. 

\subsection{Training MeMOT}
\label{sec:method:training}

We supervise MeMOT with the \textit{tracking loss} computed on the $o_i^t$, $u_i^t$, $\mathbf{b}_i^t$ following the assignment process above as
\begin{equation}
%\small
    L_{tck} = \lambda_{cls}(L'_{obj}+L'_{uni})+\lambda_{L_1}L'_{bbox}+\lambda_{iou}L'_{iou},
\end{equation}
where $\lambda$s are hyper-parameters for weight scaling, $L_{obj}$ and $L_{uni}$ are focal loss on objectness scores and uniqueness scores, $L_{bbox}$ is L1 loss for bounding box regression, $L_{iou}$ is the generalized IoU loss~\cite{rezatofighi2019generalized}.

Additionally, we apply a detection loss to the proposal embedding similar to Deformable DETR's~\cite{zhu2020deformable} to enhance MeMOT's localization capability.
Specifically, we attach an auxiliary linear decoder to the proposal embedding to output bounding boxes and object classification scores. We then assign the object instances to them as in normal object detection tasks~\cite{zhu2020deformable} and similarly compute the loss as
\begin{equation}
%\small
   L_{det} = \lambda_{cls}L_{obj}+\lambda_{L_1}L_{bbox}+\lambda_{iou}L_{iou}.
\end{equation}
Note the auxiliary decoder is discarded after training. 

% , which is matched by the Hungarian algorithm, while those for the $\Theta_D$'s are determined by object identities (\ie, each query is responsible for a track).
% Formally, given the prediction results from the $\Theta_H$ decoder and $\Theta_D$ as $\hat{Y}_\tau$ and $\hat{Y'}_\tau$, respectively, their \textit{matched} ground truths are $Y_\tau$ and $Y'_\tau$ (depending on whether the object identity is considered), the losses are formulated as:
% %
% \begin{equation}
% \footnotesize
%     \begin{split}
%     L_{det}(Y_i,\hat{Y}_i) = &\lambda_{cls}L_{obj}+\lambda_{L_1}L_{bbox}+\lambda_{iou}L_{iou}, \\ 
%      L_{tck}(Y'_i,\hat{Y'}_i) = & \lambda_{cls}(L'_{obj}+L'_{uni})+\lambda_{L_1}L'_{bbox}+\lambda_{iou}L'_{iou},
%     \end{split}
% \end{equation}
% %
% where $\lambda$s are hyper-parameters for weight scaling, $L_{obj}$ and $L_{uni}$ are focal loss on objectness scores and uniqueness scores, $L_{bbox}$ is L1 loss for bounding box regression, $L_{iou}$ is the generalized IoU loss~\cite{rezatofighi2019generalized}.

%$L_{cls}$ is the objectness score $\hat{Y}_i$ and confidence score for $\hat{Y'}_i$ (defined in Eq.~\ref{eq:score}).

Following MOTR~\cite{zeng2021motr}, we compute the tracking loss in a clip by the sum of all individual track queries' losses normalized by the total number of object instances.
For a clip with $T$ frames, the overall loss $L_{clip}$ is a combination of the tracking loss and auxiliary detection loss as:
\begin{equation}
    \footnotesize
    \begin{split}
    L_{clip} &= \lambda_{tck}L_{clip-tck}+\lambda_{det}L_{clip-det} \\
    &=\frac{\lambda_{tck}}{\sum_{t=0}^{T} N_t}\sum_{t=0}^T\sum_{i=0}^{|\mathbi{Q}^t_{tc}, \mathbi{Q}^t_{pro}|}{L_{tck}^{(i, t)}}+\frac{\lambda_{det}}{T}\sum^T_{t=0}\frac{1}{N_t}\sum_{j=0}^{|\mathbi{Q}^t_{pro}|}L_{det}^{(j, t)},
    \end{split}
    \label{eq:cliploss}
\end{equation}
where $\lambda_{tck} \in \mathbb{R}$ and $\lambda_{det} \in \mathbb{R}$ are the loss weights for balancing the tracking loss and the auxiliary detection loss, respectively. 
Here $N_t$ denotes the total number of visible objects in the frame at time $t$.

%(\ie, new objects plus tracked objects) of bounding boxes in frame $t$.
\section{Experiments}
\label{sec:exp}

\begin{table*}[t!]
    \centering
    \footnotesize
    \setlength\tabcolsep{4.0pt}
    \begin{tabular}{l|ccccccccccc}
    \toprule[1.5pt]
        \textbf{Method} & \textbf{Training Data} & \textbf{Transformer} &\textbf{IDF1}  $\uparrow$ & \textbf{MOTA} $\uparrow$ & \textbf{HOTA} $\uparrow$ & \textbf{AssA} $\uparrow$ & \textbf{IDsw} $\downarrow$ & \textbf{MT(\%)} $\uparrow$ & \textbf{ML(\%)} $\downarrow$ & \textbf{FP} $\downarrow$ & \textbf{FN} $\downarrow$  \\\hline
        \multicolumn{12}{c}{\textbf{MOT16} \cite{milan2016mot16}} \\\hline\hline
        FairMOT~\cite{zhang2020fair} & 13.1x & & 72.3 & 69.3 & 58.3 & 58.0 & 815& 40.3& 16.7 & 13501 & 41653\\
        TubeTK~\cite{pang2020tubetk} & 44.5x & & 62.2 & 66.9 & 50.8 & 47.3 & 1236 & 39.0 & 16.1 & 11544 & 47502 \\
        CTracker~\cite{peng2020chained} & 1.0x & & 57.2 & 67.6 & 48.8 & 43.7 & 1897 & 32.9 & 23.1 & 8934 & 48350 \\ 
        JDE~\cite{wang2019towards} & 10.2x & & 55.8 & 64.4 & - & - & 1544 & 35.4 & 20.0 & - & - \\
        \rowcolor{gray!20} MOTR \cite{zeng2021motr} & 1.9x & \checkmark &  67.0 & 66.8 & -& -& \textbf{586} & 34.1 & 25.7 & \textbf{10364} & 49582 \\
        \rowcolor{gray!20} \textbf{MeMOT} (ours) & 1.9x & \checkmark & \textbf{69.7} & \textbf{72.6} & \textbf{57.4} & \textbf{55.7} & 845 & \textbf{44.9} & \textbf{16.6} & 14595 & \textbf{34595} \\\hline
        \multicolumn{12}{c}{\textbf{MOT17} \cite{milan2016mot16}} \\\hline\hline
        CorrTracker~\cite{wang2021multiple} & 13.1x & & 73.6 & 76.5 & 60.7 & 58.9 & 3396 & 47.6 & 12.7 & 29808 & 99510\\
        FairMOT \cite{zhang2020fair} & 13.1x & & 72.3 & 73.7 & 59.3 & 58.0 & 3303 & 43.2 & 17.3 & 27507 & 117477\\
        PermaTrack~\cite{tokmakov2021learning} & 18.7x & & 68.9 & 73.8 & 55.5 & 53.1 & 3699 & 43.8 & 17.2 & 28998 & 115104 \\
        GSDT~\cite{wang2021joint} & 10.2x &&  66.5 & 73.2 & 55.2 & 51.0 & 3891 & 41.7 & 17.5 & 263397 & 120666 \\
        TraDeS~\cite{wu2021track} & 3.8x & &  63.9 & 69.1 & 52.7 & 50.8 & 3555 & 36.4 & 21.5 & 20892 & 150060 \\
        TransTrack~\cite{sun2020transtrack} & 3.8x & \checkmark &  63.5 & 75.2 & 54.1 & 47.9 & 4614 & 55.3 & 10.2 & 50157 & 86442 \\
        TransCenter~\cite{xu2021transcenter} & 3.8x & \checkmark & 62.2 & 73.2 & 54.5 & 49.7 & 3663 & 40.8 & 18.5 & 23112 & 123738 \\
        TubeTK~\cite{pang2020tubetk} & 44.5x & & 58.6 & 63.0 & 48.0 & 45.1 & 4137 & 31.2 & 19.9 & 27060 & 177483 \\
        CTracker~\cite{peng2020chained} & 1.0x & & 57.4 & 66.6 & 49.0 & 37.8 & 5529 & 32.2 & 24.2 & 22284 & 160491 \\
        \rowcolor{gray!20} TrackFormer~\cite{meinhardt2021trackformer} & 1.0x & \checkmark & 63.9 & 65.0 & - & -& 3258 & - & - & 70443 & 123552 \\
        \rowcolor{gray!20} MOTR~\cite{zeng2021motr} & 1.9x & \checkmark & 67.0 & 67.4& - & - & \textbf{1992} & 34.6 & 21.5 & \textbf{32355} & 149400 \\
        \rowcolor{gray!20} \textbf{MeMOT} (ours) & 1.9x & \checkmark & \textbf{69.0} & \textbf{72.5} & \textbf{56.9} & \textbf{55.2} & 2724 & \textbf{43.8} & \textbf{18.0} & 37221 & \textbf{115248} \\ \hline
        \multicolumn{12}{c}{\textbf{MOT20} \cite{dendorfer2020mot20}} \\\hline\hline
        FairMOT~\cite{zhang2020fair} & 8.2x & & 67.3 & 61.8 & 54.6 & 54.7 & 5243 & 68.8 & 7.6 & 103440 & 88901\\
        TransTrack~\cite{sun2020transtrack} & 2.7x & \checkmark &  59.4 & 65.0 & 48.9 & 45.2 & 3608 & 50.1 & 13.4 & 27191 & 150197 \\
        TransCenter~\cite{xu2021transcenter} & 2.7x & \checkmark & 49.6 & 58.5 & 43.5 & 37.0 & 4695 & 48.6 & 14.9 & 64217 & 146019 \\
        \rowcolor{gray!20} \textbf{MeMOT} (ours) & 1.0x & \checkmark & \textbf{66.1} & \textbf{63.7} & \textbf{54.1} & \textbf{55.0} & \textbf{1938} & \textbf{57.5} & \textbf{14.3} & \textbf{47882} & \textbf{137983} \\
    \bottomrule[1.5pt]
    \end{tabular}
    \vspace{-2.0mm}
    \caption{\textbf{Evaluation results on MOT challenge datasets}. Trackers with gray background use the in-network association solver (IAS), and others with white background use the post-model association solver (PAS).
    Best results of IAS are marked in bold.}
    \label{tab:MOTexp}
    \vspace{-1.5mm}
\end{table*}

\subsection{Datasets and Metrics} 
\label{sec:exp:datasets}

We evaluate MeMOT on \textbf{MOT Challenge}~\cite{milan2016mot16,dendorfer2020mot20} (\ie, MOT16, 17~\&~20) datasets.
As standard protocols, \textbf{CLEAR MOT Metrics}~\cite{milan2016mot16} and \textbf{HOTA}~\cite{luiten2021hota} are used for evaluation.

\subsection{Settings}
\label{sec:exp:settings}

\noindent \textbf{Implementation Details}.
We implemented our proposed method in PyTorch~\cite{paszke2019pytorch}, and performed all
experiments on a system with 8 Tesla A100 GPUs.
The input frames were resized such that their shorter side is 800 pixels.
We used routine data augmentations, including random flip and crop.
We adopted ResNet50~\cite{he2016deep} and Deformable DETR~\cite{zhu2020deformable} pretrained on COCO~\cite{lin2014microsoft} for hypothesis generation.
For all Transformer units, we reduced their number of layers to 4.
Our memory buffer contained a maximum of 300 tracks for MOT16/17 benchmarks and 600 tracks for MOT20. Its maximum temporal length is 22 for MOT16/17 and 20 for MOT20, which is mainly limited by the GPU memory.
We followed prior work~\cite{carion2020end,zhu2020deformable} and selected the coefficients of Hungarian loss with $\lambda_{cls}$, $\lambda_{L_1}$ and $\lambda_{iou}$ as 2, 5, 2, respectively.
We set $\lambda_{det}=\lambda_{tck}=1$ in Eq.~\ref{eq:cliploss}.

\vspace{3pt} \noindent \textbf{Hyperparameters}.
We adopted clip-centric training.
The length of each clip started from 2 and increased with stride 4 at every 20 epochs.
Frames in each clip were sampled with a random interval between 1 to 10.
Our model was trained with AdamW~\cite{loshchilov2017decoupled} optimizer for 200 epochs.
The learning rate was initiated to $2\times10^{-4}$ and decreased by 10 at the 100-th epoch.
The batch size was set to 1 clip per GPU.

\vspace{3pt} \noindent \textbf{Training Data}.
When compared with the state-of-the-art methods, MeMOT is trained on the CrowdHuman~\cite{shao2018crowdhuman} validation set and MOT17 training set for MOT16 and MOT17 benchmarks.
No extra data was used for MOT20.
Training with extra data can significantly boost the tracking performance~\cite{zhang2020fair}.
Thus, as shown in Table~\ref{tab:MOTexp}, we mark out the size of additional training data (\ie, number of frames) that each method used, where the MOT training set itself is referred to as $1.0\times$.
More details are included in the appendix.

\subsection{Comparison with the State-of-the-art Methods}
\label{sec:exp:sota}

For fair comparison, we mainly compare MeMOT to methods with an \textit{in-network association solver (IAS)} that predict identities directly without any post-processing. The other type of method applies a \textit{post-network association solver (PAS)} on detection results to perform a series of rule-based linking, such as Hungarian matching with Kalman Filter and re-ID features.
Generally, these empirical linking strategies limit their practicability and scalability.

\vspace{3pt} \noindent \textbf{Results on normal scenarios}.
Table~\ref{tab:MOTexp} shows that MeMOT achieves the state-of-the-art performance in MOT16/17 among the IAS methods (w/ gray background).
It also obtains encouraging detection accuracy (72.6 and 72.5 MOTA on MOT16/17) compared to the PAS methods that are pretrained with larger detection datasets.
For more comprehensive metrics, IDF1 and HOTA, MeMOT achieves comparable results with the state-of-the-art JDE tracker (FairMOT), but uses 5$\times$ less training data.
MeMOT can keep track of more objects but produce much fewer ID switches (IDsw).
For example, on MOT16, MeMOT obtains 44.9\% Mostly Tracked (MT) and 16.6\% Mostly Lost (ML), outperforming other methods by at least 4.5\%, but only getting 845 IDsw.
On MOT17, TransTrack and TransCenter show promising detection results with better MT (55.3) and ML (10.2), however, they produce 34\% and 69\% more IDsw and lower IDF1 (63.5 vs. 62.2 vs. 69.0) than ours.
Compared to all Transformer-based methods, MeMOT is significantly better in data association measured by Association Accuracy (AssA).
This shows the effectiveness of the learnable association powered by our memory design.

\begin{figure}[t]
    \centering
    \includegraphics[width=0.99\linewidth]{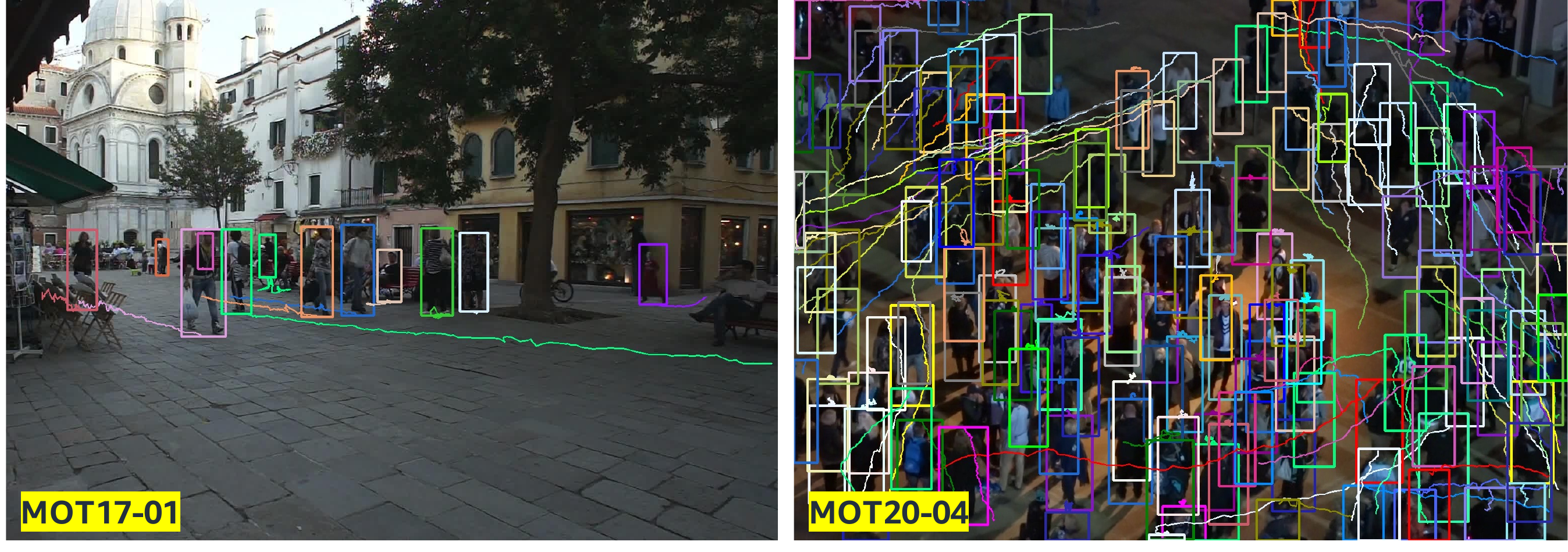}
    \vspace{-1.5mm}
    \caption{
        \textbf{Examples of our tracking performance} on MOT17 and MOT20. Each identity is shown in a colored bounding box and trajectory in the past 150 frames are displayed.}
    \label{fig:vis_result}
    \vspace{-3.0mm}
\end{figure}

\vspace{-4pt}
\vspace{3pt} \noindent \textbf{Results on crowded scenarios}.
MOT20 is a more challenging benchmark with crowded scenarios and serious occlusions.
Table~\ref{tab:MOTexp} shows that MeMOT achieves comparable performance with the state-of-the-art JDE method (FairMOT), but gets 63\% reduction in IDsw.
Note that FairMOT is trained with 8$\times$ more training data than ours.
Comparing to other Transformer-based methods, MeMOT outperforms them by 6.7 IDF1 and 5.2 HOTA.
By getting a much lower IDsw, our learnable association solver shows its advantage to deal with the occlusion problem.
We observe that IoU-based association methods (\eg, TransCenter and TransTrack)
fail to handle frequent occlusion, and for the re-identification feature-based methods (\eg, FairMOT), it is hard to obtain high-quality embeddings to measure inter-object similarity due to the small object sizes.

\begin{figure}
\centering
    \includegraphics[width=\linewidth]{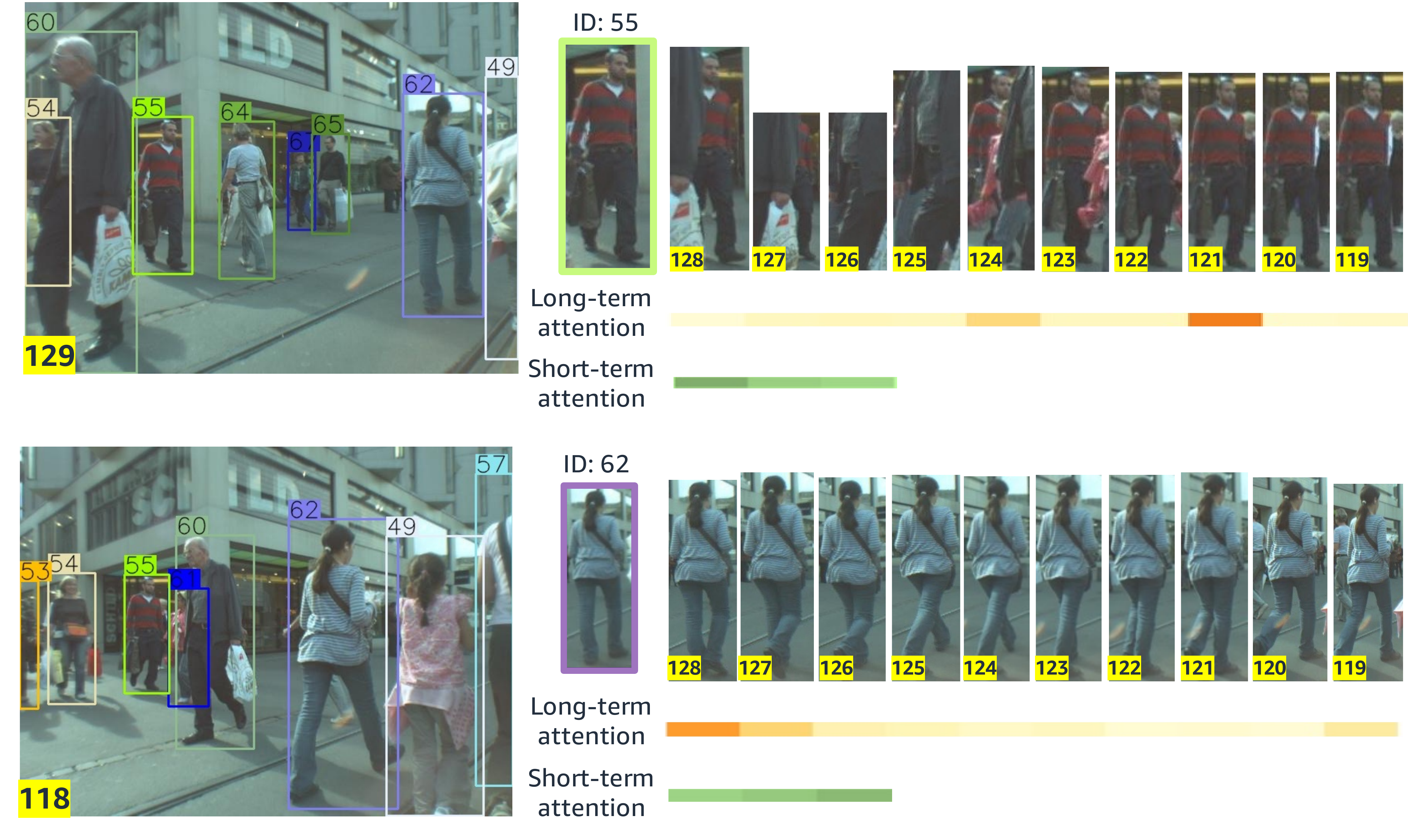}
    \vspace{-6.0mm}
    \caption{
    \textbf{Visualization of long- and short-term attentions}.
    \textbf{Left}: Tracking results of frame 118 and 129, where tracked objects are displayed in colors with confidence scores.
    \textbf{Right}: Learned long- and short-term attention maps for the intermediate frames of two selected identities (\ie, ID 55 and 62). Darker color represents stronger attention.}
    \label{fig:vis_attn}
    \vspace{-2.5mm}
\end{figure}

\vspace{-1mm}
\subsection{Visualization}
\label{sec:exp:qualitative}
\vspace{-1mm}

Object trajectories are visualized in Fig.~\ref{fig:vis_result}. Results of MOT17-01 show that MeMOT generates long, consistent predictions even when objects pass by each other frequently. Results on MOT20-04 suggest MeMOT's superior object detection and association ability in crowd scenarios. Due to the small object size and poor lighting, feature similarity-based association methods~\cite{zhang2020fair,wang2019towards} are precarious, causing higher IDsw. We provide video demos in the supplementary material for detailed comparison.

In Fig.~\ref{fig:vis_attn}, we also visualize the attention weights of the memory aggregator to elaborate what information is referenced from the memory. For object 55, who is occluded by object 60 from frame 125 to 128, the embedding right before occlusion (frame 124) and a full-body feature (frame 121) contribute the most to re-linking (frame 129) after occlusion. And his short-term attention weights are higher on a less-occluded frame (frame 128) than fully-occluded frames (frame 126 and 127).
As for a non-occluded object (\ie, object 62), attention weights are higher on the short-term memory (frame 126 to 128) and the far-away frames are less attended. These observations validate that our memory aggregator is capable of capturing distinctive object features, especially when objects are crossing each other.

\subsection{Ablation Studies}
\label{sec:exp:ablation}

We experiment with different memory and model design choices.
Unless noted otherwise, we use trimmed models by reducing their number of layers of all Transformer units from 4 to 2.
% The longer side of input frames is reduced to $1333$ pixels.
The models are trained on MOT17 training set and validated on MOT15 training set. Validation videos that are overlapped with the training set are excluded.

\vspace{3pt} \noindent \textbf{Effect of short-term memory length}.
Table~\ref{tab:ablation:memory_length_short} compares the performance using different short-term memory lengths, by keeping the long-term memory length $T_l$ as 24.
It shows that only using the last two observations (\ie, $T_s$=$2$) for short-term memory aggregation slightly decreases the performance.
This observation is consistent with results in prior work~\cite{sun2020transtrack,meinhardt2021trackformer} that propagates tracking results only between adjacent frames.
On the other hand, increasing the short-term length from 3 to 5 does not make a big difference.
We think these information gaps are compensated by the long-term memory.
Considering the accuracy-efficiency trade-off, we set $T_s$ to 3 as default in other experiments.    

\begin{table}
\centering
\footnotesize
\begin{tabular}{cc|cccccc}
\toprule[1.5pt]
        \textbf{$T_s$} & \textbf{$T_l$} & \textbf{IDF1} & \textbf{MOTA} & \textbf{HOTA} & \textbf{IDsw} & \textbf{DetA} & \textbf{AssA} \\\hline
         2 & \multirow{4}{*}{24} & 72.52& 65.62& 58.99& \textbf{76}& 56.97& 62.14\\
        3 & &\textbf{73.15}& \textbf{68.08}& \textbf{59.75} & 93 & \textbf{57.92} & \textbf{63.10}\\
        4 & &72.40 & 67.11& 59.51 & 93& 57.58& 62.82\\
        5 & &72.75 & 66.65 & 59.48& 92& 57.42& 62.87\\
\bottomrule[1.5pt]
    \end{tabular}
\vspace{-2.0mm}
\caption{Comparisons on different length of short-term memory.}
\vspace{-2.0mm}
\label{tab:ablation:memory_length_short}
\end{table}
\begin{table}
\centering
\footnotesize
\begin{tabular}{cc|cccccc}
\toprule[1.5pt]
        \textbf{$T_s$} & \textbf{$T_l$} & \textbf{IDF1} & \textbf{MOTA} & \textbf{HOTA} & \textbf{IDsw} & \textbf{DetA} & \textbf{AssA} \\\hline
        \multirow{5}{*}{3} & 3 & 71.27 & 67.14 & 59.09 & 136 &57.85& 61.41\\
        & 5 & 71.70 & 67.94 & 59.31 & 136 & 58.24 & 61.50\\
        & 10 & 71.66 & \textbf{68.29}& 59.53 & 117 & \textbf{58.35} & 59.49\\
        & 20 & 72.83&  68.21 & \textbf{59.85} & 96& 58.03 & 63.00\\
        & 24 & \textbf{73.15}& 68.08& 59.75 & \textbf{93} & 57.92 & \textbf{63.10}\\
\bottomrule[1.5pt]
    \end{tabular}
\vspace{-2.0mm}
\caption{Comparisons on different length of long-term memory.}
\label{tab:ablation:memory_length_long}
\vspace{-2.0mm}
\end{table}
\begin{table}
\centering
\footnotesize
\begin{tabular}{cc|cccccc}
\toprule[1.5pt]
        \textbf{$q_s$}  & \textbf{$q_l$} & \textbf{IDF1} & \textbf{MOTA} & \textbf{HOTA} & \textbf{IDsw} & \textbf{DetA} & \textbf{AssA} \\\hline
        & \checkmark  & \textbf{73.15}& 68.08 & \textbf{59.75}& \textbf{93}& 57.92 & 63.10\\
        \checkmark  & & 67.25 & \textbf{69.88} & 57.36 & 112 & \textbf{58.01} & 61.76\\
        \checkmark  & \checkmark & 41.09& 59.80& 43.28& 207 & 45.36 & 38.52\\
        & & 72.30 & 62.68 & 58.84 & 103 & 55.72 & \textbf{63.37}\\
\bottomrule[1.5pt]
    \end{tabular}
    \vspace{-2.0mm}
\caption{Comparisons on different configuration of the short-term cross-attention query $q_s$ and long-term cross-attention query $q_l$.}
%\caption{Comparisons on different configuration of the short-term cross-attention query $q_s$ and long-term cross-attention query $q_l$. Empty means using the latest observation, $\checkmark$ represents the query is learnable and recursively updated during inference. Our MeMoT design (row-1) yields best overall performance. }
\label{tab:ablation:learn}
\vspace{-2.0mm}
\end{table}

\vspace{3pt} \noindent \textbf{Effect of long-term memory length}.
MeMOT uses a long-term memory to mitigate the occlusion problem.
Table~\ref{tab:ablation:memory_length_long} shows the effect of different long-term memory lengths $T_l$ from 3 to 24.
Note that we set the max length as 24 due to hardware limitation.
As $T_l$ grows, the association performance keeps increasing with fewer IDsw and higher IDF1.

\vspace{3pt} \noindent \textbf{Comparing to heuristic memory aggregations}.
We explore the design of memory aggregation module by first comparing to heuristic algorithms.
Considering the tracklet length can be relatively long (up to 24),
we do not concatenate the embeddings but test the pooling methods.
Then the aggregation can be conducted by using either arithmetic mean or maximum norm over the most recent $T$ frames.
Table~\ref{tab:ablation:mem_structure_heuristic} shows that using these simple pooling methods is incapable of capturing informative track features, resulting in a huge performance drop in IDF1 and MOTA.

\vspace{3pt} \noindent \textbf{Comparing to attention-based memory aggregations}.
We experiment with another two attention-based aggregation designs in memory encoding.
The first one is to only use a cross-attention module, without the separation of long and short memory.
This baseline uses the latest observation to query an object's past $T$ embeddings.
As shown in Table~\ref{tab:ablation:mem_structure_attn}, it produces worse association performance, with -0.51\% and -0.34\% MOTA for $T=3$ and $T=24$, respectively.
The IDsw also increases by 6 and 44.
Inspired by LSTR~\cite{xu2021long}, the second one is to use the aggregated short-term embeddings to retrieve useful information from the long-term memory. The result shows that this design also decreases the performance. We argue that, in the action detection task that LSTR focuses on, the result of each frame is independent and deficient short-term features have a limited effect on future predictions. However, association errors can propagate in MOT, thus using long-term features to compensate for short-term features is more desirable. 

\begin{table}
\centering
\footnotesize
\begin{tabular}{c|c|cccc}
\toprule[1.5pt]
        \textbf{Method} & \textbf{Parameter} &\textbf{IDF1} & \textbf{MOTA} & \textbf{HOTA} & \textbf{IDs} \\\hline
        Ours & - &\textbf{73.15}& \textbf{68.08}& \textbf{59.75}& \textbf{93}\\\hline
        \multirow{4}{*}{Pooling}& Average (T=3) & 25.04& 30.72 & 21.89 & 267\\
        & Max (T=3) & 46.83& 41.28& 35.44& 235\\ 
        & Average (T=24) & -& -7.29& -& -\\
        & Max (T=24) & 25.78& 10.20& 6.54& 332\\
\bottomrule[1.5pt]
    \end{tabular}
    \vspace{-2.0mm}
\caption{Comparisons with heuristic memory aggregation design.}
%\caption{Comparisons with heuristic memory aggregation design. The result shows simple heuristic aggregation methods are incapable of extracting instance features for tracking, even given the long-term memory. Our proposed memory aggregator yields significantly better tracking performance.}
\label{tab:ablation:mem_structure_heuristic}
\vspace{-2.0mm}
\end{table}
\begin{table}
\footnotesize
\begin{tabular}{c|c|cccc}
\toprule[1.5pt]
        \textbf{Method} & \textbf{Parameter} &\textbf{IDF1} & \textbf{MOTA} & \textbf{HOTA} & \textbf{IDs} \\\hline
        Ours & - &\textbf{73.15}& 68.08& \textbf{59.75}& \textbf{93}\\\hline
        \multirow{2}{*}{Single}& T=3 & 72.64 & \textbf{68.74} & 58.94 & 137\\
        & T=24 & 72.81 & 66.25 & 58.73 & 99\\\hline
        \makecell{Long-after-short} & - & 70.30 & 65.39& 57.03 & 101\\
\bottomrule[1.5pt]
\end{tabular}
\vspace{-2.0mm}
\caption{Comparisons on adaptive memory aggregation design.}
\label{tab:ablation:mem_structure_attn}
\vspace{-2.0mm}
\end{table}
\begin{table}[t]
\centering
\footnotesize
\begin{tabular}{c|cccccc}
\toprule[1.5pt]
        \textbf{Update}  & \textbf{IDF1} & \textbf{MOTA} & \textbf{HOTA} & \textbf{IDsw} & \textbf{DetA} & \textbf{AssA} \\\hline
        \checkmark  & \textbf{73.15}& \textbf{68.08}& \textbf{59.75} & \textbf{93} & \textbf{57.92} & \textbf{63.10}\\
         & 61.03 & 43.42 & 49.24 & 161 & 42.40 & 57.96 \\
\bottomrule[1.5pt]
    \end{tabular}
    \vspace{-2.0mm}
\caption{Comparison on the updating of $\mathbi{Q}_{dmat}$.}
\label{tab:ablation:update}
\vspace{-2.0mm}
\end{table}
\begin{table}
\footnotesize
\begin{tabular}{c|cccccc}
\toprule[1.5pt]
        \textbf{Confidence} & \textbf{IDF1} & \textbf{MOTA} & \textbf{HOTA} & \textbf{IDs}& \textbf{DetA} & \textbf{AssA} \\\hline
        Single& 69.09& 63.51& 52.86& 104& 55.76 & 61.77\\
        Dual & \textbf{73.15}& \textbf{68.08}& \textbf{59.75} & \textbf{93} & \textbf{57.92} & \textbf{63.10}\\
\bottomrule[1.5pt]
    \end{tabular}
\vspace{-2.0mm}
\caption{Comparisons between single and dual confident scores.}
\label{tab:ablation:novelty}
\vspace{-2.0mm}
\end{table}

\vspace{3pt} \noindent \textbf{Using learnable tokens vs. latest observation for memory aggregation}.
We explore using either the learnable tokens or the latest observation for long- and short-term memory aggregation, as shown in Table~\ref{tab:ablation:learn}.
For the short-term token (row2 vs. row4), using the latest observation (row4) yields better association performance (+5.05 IDF1).
After fixing the short-term token, using learnable tokens for long-term memory aggregation obtains slightly better performance (row1 vs. row4), with +0.85 IDF1 and -10 IDsw.
It is worth noting that using learnable tokens for both long- and short-term branches is risky, getting the IDF1 and MOTA drop to 41.09\% and 59.80\%.
These observations validate our intuition for separating the long- and short-term branches.
1) Due to the temporal variance, the long-term memory may be less informative to match the latest observation but provides diverse features within a tracklet.
To extract the supportive context information, using learnable tokens is more effective.
2) Short-term memory features share a high similarity, thus directly querying them with the latest observation can smooth out the noises.
3) These two branches can acquire complementary information.

\vspace{3pt} \noindent \textbf{Dynamic update of memory aggregation tokens}.
Since we model online tracking as an iterative process, it is worth studying if the long-term memory aggregation tokens should be updated during inference.
Results in Table~\ref{tab:ablation:update} shows that dynamically updating the queries with the most recent information contributes to better detection and tracking performance. By passing the firsthand observation to the long-term queries, more detailed information for current association is extracted, rather than general information.

\vspace{3pt} \noindent \textbf{Effect of uniqueness score for training MeMOT}.
We introduce the uniqueness score to link new detections with the tracked objects and reject false positives.
Here we evaluate its contribution by removing the prediction branch of uniqueness score, as shown in Table~\ref{tab:ablation:novelty}.
Without the uniqueness branch (single output), there are more false positive detections and IDsw.
We separate the mixed meanings of the output classification score by splitting the prediction of objectness and uniqueness into two heads. In a single-head architecture, for tracked object queries, the score means the confidence of existence; for the proposal queries, it means the confidence of being a new-born object. While the two purposes share the same classification layer, low confidence values are ambiguous: it means non-objectness, or not a new object. Our design removes the ambiguity and avoids under-training of the classification layers.

\subsection{Limitations}
\label{sec:exp:limitations}

As MeMOT is currently trained with supervised learning, it requires video datasets with tracking annotation. 
However, existing datasets for tracking are still limited in size and diversities, due to the high cost of annotating videos. 
Developing annotation efficient training methods is crucial to overcoming this difficulty.
Although the spatio-temporal memory is shown to be effective in tracking objects consistently, it indeed increases the GPU memory cost in training. This limits the temporal length of the memory and therefore calls for further improvement in efficiency.

\vspace{-1mm}
\section{Conclusion}
\label{sec:conclusion}
\vspace{-1mm}

We proposed MeMOT for online MOT by jointly performing the object detection and data association.
MeMOT preserves a large spatio-temporal memory and actively encodes the past observations via an attention-based aggregator. 
By representing objects as dynamically updated query embeddings, MeMOT predicts object states with an attention mechanism without any post-processing. Extensive experiments validate the effectiveness of MeMOT on object localization and association in crowded scenes.

There are many real-world applications of MOT technology, such as patient or elderly health monitoring, autonomous driving, and collaborative robots. However, there could be unintended usages and we advocate responsible usage complying with applicable laws and regulations.

\vspace{-0.5mm}
\section{Appendix}

\subsection{Algorithm}
\vspace{-2.5mm}
\begin{algorithm}
    \small
    \caption{MeMOT Algorithm}
    \hspace*{\algorithmicindent} \textbf{Input}: A sequence of video frames $\mathbi{I} = \{I^0, I^1, \cdots, I^T\}$.\\
    \hspace*{\algorithmicindent} \textbf{Memory}: A set of track states $\mathbi{X} = \{X_0, X_1, \cdots, X_K\}$ with $X_k = \{\widehat{\mathbi{q}}_k^{t_0}, \widehat{\mathbi{q}}_k^{t_1}, \cdots, \widehat{\mathbi{q}}_k^{t_N}\}$ of identity embeddings $\widehat{\mathbi{q}}_k^t \in \mathbb{R}^d$.\\
    % $\widehat{\mathbi{q}}_k^t \leftarrow \mathbi{0}$ if object $k$ does not appear in $I^t$. \\
    \hspace*{\algorithmicindent} \textbf{Output}: A set of trajectories $\mathbfcal{T} = \{\mathcal{T}_0, \mathcal{T}_1, \cdots, \mathcal{T}_K\}$ with $\mathcal{T}_k = \{\mathbi{b}_k^{t_0}, \mathbi{b}_k^{t_1}, \cdots, \mathbi{b}_k^{t_N}\}$ of bounding boxes $\mathbi{b}_k^{t} = (x, y, w, h)$.
    \begin{algorithmic}[1]
    \State $\mathbi{X} \leftarrow \O$
    \For{$I^t \in \mathbi{I}$}
        \State $\mathbfcal{T}_{new}, \mathbi{X}_{new} \leftarrow \O, \O$
        
        % hypothesis generation
        \State $z_1^t \leftarrow \Theta_H$.encoder($I^t$)
        \State $\mathbi{Q}_{pro}^t \leftarrow \Theta_H$.decoder($z_1^t$)
        
        % encoding
        \State $\mathbi{Q}^t_{AST} \leftarrow \Theta_E$.$f_{short}(\mathbi{X}^{t-1}, \mathbi{X}^{t-1-T_s:t-1})$
        \State $\mathbi{Q}^t_{ALT} \leftarrow \Theta_E$.$f_{long}(\mathbi{Q}_{dmat}^{t-1}, \mathbi{X}^{t-1-T_l:t-1})$
        \State $[\mathbi{Q}_{tck}^t, \mathbi{Q}_{dmat}^{t}] \leftarrow \Theta_E$.$f_{fuse}([\mathbi{Q}^t_{AST}, \mathbi{Q}^t_{ALT}])$
        
        % decoding
        \State $[\widehat{\mathbi{Q}}_{pro}^t, \widehat{\mathbi{Q}}_{tck}^t] \leftarrow \Theta_D$.association\_solver$([\mathbi{Q}_{pro}^t, \mathbi{Q}_{tck}^{t}], z_1^t)$
        \State $[\mathbi{B}^t_{pro}, \mathbi{B}^t_{tck}], [\mathbi{S}^t_{pro}, \mathbi{S}^t_{tck}] \leftarrow \Theta_D$.predictor$([\widehat{\mathbi{Q}}_{pro}^t, \widehat{\mathbi{Q}}_{tck}^t])$
        
        % update
        \For{$\mathbi{b}^t_{k} \in \mathbi{B}^t_{tck}$, $\mathbi{s}^t_{k} \in \mathbi{S}^t_{tck}$, $\widehat{\mathbi{q}}^t_{k} \in \widehat{\mathbi{Q}}^t_{tck}$}
            \If{$\mathbi{s}^t_{k} \ge$ $\epsilon_{tck}$}
                \State $\mathcal{T}_k \leftarrow \mathcal{T}_k + \{\mathbi{b}^t_{k}\}$
                \State $X_k \leftarrow X_k + \{\widehat{\mathbi{q}}^t_{k}\}$
            \EndIf
        \EndFor
        \For{$\mathbi{b}^t_{k} \in \mathbi{B}^t_{pro}$, $\mathbi{s}^t_{k} \in \mathbi{S}^t_{pro}$, $\widehat{\mathbi{q}}^t_{k} \in \widehat{\mathbi{Q}}^t_{pro}$}
            \If{$\mathbi{s}^t_{k} \ge$ $\epsilon_{pro}$}
                \State $\mathbfcal{T}_{new} \leftarrow \mathbfcal{T}_{new} + \{\{\mathbi{b}^t_{k}\}\}$
                \State $\mathbi{X}_{new} \leftarrow \mathbi{X}_{new} + \{\{\widehat{\mathbi{q}}^t_{k}\}\}$
            \EndIf
        \EndFor
        \State $\mathbfcal{T} \leftarrow \mathbfcal{T} + \mathbfcal{T}_{new}$
        \State $\mathbi{X} \leftarrow \mathbi{X} + \mathbi{X}_{new}$
    \EndFor
    \end{algorithmic}
    \label{alg:memot}
\end{algorithm}

The workflow of our proposed MeMOT is shown in Algorithm~\ref{alg:memot}. 
MeMOT takes a sequence of video frames $\mathbi{I} = \{I^0, I^1, \cdots, I^T\}$ as input, and outputs trajectories $\mathbfcal{T} = \{\mathcal{T}_0, \mathcal{T}_1, \cdots, \mathcal{T}_K\}$ for $K$ objects. The track states $\mathbi{X} = \{X_0, X_1, \cdots, X_K\}$, represented as embeddings for each object at its own active timestamps, are maintained and updated in a spatio-temporal memory buffer. 
MeMOT contains three Transformer-based network modules: 1) a hypothesis generation module $\Theta_H$ for extracting the frame feature $z_1$ and producing the proposal embeddings $\mathbi{q}_{pro}$, 2) a memory encoding module $\Theta_E$ that aggregates the previous states to track embeddings $\mathbi{q}_{tck}$ for each object, and 3) a memory decoding module $\Theta_D$ that predicts the current states of tracked objects and initializes new objects. 

Concretely, at time $t$, the encoder of $\Theta_H$ translates image $I^t$ to features $z_1^t \in \mathbb{R}^{d \times HW}$, which are then decoded to a set of proposal embeddings $\mathbi{Q}_{pro}^t$ by $\Theta_H$'s decoder.
At the same time, the short-term aggregation module $f_{short}$ in $\Theta_E$ queries the past $T_s$ memory $\mathbi{X}^{t-1-T_s:t-1}$ with the latest observation $\mathbi{X}^{t-1}$ and obtains the aggregated short-term queries $\mathbi{Q}^t_{AST}$.
The long-term aggregation module $f_{long}$ uses a set of learnable queries, called dynamic memory aggregation token (DMAT) $\mathbi{Q}_{dmat}^{t-1}$, and takes advantages of a longer time period $T_l$ to produce the aggregated long-term queries $\mathbi{Q}^t_{ALT}$. $\mathbi{Q}^t_{AST}$ and $\mathbi{Q}^t_{ALT}$ are fused by a self-attention module $f_{fuse}$, which outputs the track query $\mathbi{Q}_{tck}^t$ and updated $\mathbi{Q}_{dmat}^{t}$. 
$\Theta_D$ takes the concatenated set of $\mathbi{Q}_{pro}^t$ and $\mathbi{Q}_{tck}^t$ as query set and the frame feature $z_1^t$ as key-value, generating the estimated states $\widehat{\mathbi{Q}}_{pro}^t$ and $\widehat{\mathbi{Q}}_{tck}^t$. 
Object bounding boxes $\textbf{B}_{pro}^t$, $\textbf{B}_{tck}^t$ and confidence scores $\textbf{S}_{pro}^t$, $\textbf{S}_{tck}^t$ are obtained from $\widehat{\textbf{Q}}_{pro}^t$ and $\widehat{\textbf{Q}}_{tck}^t$ through FFN. 
For all the tracked objects, the states will be updated if their confidence scores are above a threshold $\epsilon_{tck}$. Similarly, proposal queries will be initialized as new tracks if the confidence scores are higher than $\epsilon_{pro}$. 
As discussed in the paper, $T_s$ is selected as 3 as an accuracy-efficiency trade-off; while $T_l$ is 24 frames due to hardware limitation. We select $\epsilon_{pro}$, $\epsilon_{tck}$ as 0.7 and 0.6, respectively.

%%%%%%%%% REFERENCES
{\small
\bibliographystyle{ieee_fullname}
\bibliography{egbib}
}

\end{document}